%% file: arxiv.tex
\documentclass[letterpaper]{article}
\usepackage[preprint]{aaai2027}
\usepackage[hyphens]{url}
\usepackage{graphicx}
\usepackage{natbib}
\usepackage{caption}
\usepackage{booktabs}
\usepackage{amsmath}
\usepackage{amssymb}

\title{Predictive Memory Localization: Forecasting Selective Intervention Paths from Internal Signals}
\author{
    Jinhao Jing\textsuperscript{\rm 1},
    Tian Zeyu\textsuperscript{\rm 2},
    Lucas Qingyang Fang\textsuperscript{\rm 3},
    Zhisheng Chen\textsuperscript{\rm 4},
    Shuang Chen\textsuperscript{\rm 5},\\
    Yuhao Luo\textsuperscript{\rm 6}\corresponding,
    Qiannian Zhao\textsuperscript{\rm 7}\corresponding
}
\affiliations{
    \textsuperscript{\rm 1}CUHK, Shenzhen;\quad
    \textsuperscript{\rm 2}Shanghai Jiao Tong University;\quad
    \textsuperscript{\rm 3}University of California, Santa Cruz;\quad
    \textsuperscript{\rm 4}University of the Chinese Academy of Sciences;\quad
    \textsuperscript{\rm 5}University of California, Los Angeles;\quad
    \textsuperscript{\rm 6}JD.COM;\quad
    \textsuperscript{\rm 7}Kuaishou Technology
}

\begin{document}

\maketitle

\begin{abstract}
\input{sections/abstract}
\end{abstract}

\input{sections/introduction}
\input{sections/related_work}
\input{sections/problem_setup}
\input{sections/method}
\input{sections/experiments}
\input{sections/discussion}
\input{sections/limitations}
\input{sections/conclusion}

\bibliography{references}

\clearpage
\appendix
\renewcommand{\thesection}{S\arabic{section}}
\renewcommand{\thesubsection}{S\arabic{section}.\arabic{subsection}}
\renewcommand{\thetable}{S\arabic{table}}
\renewcommand{\thefigure}{S\arabic{figure}}
\renewcommand{\theequation}{S\arabic{equation}}
\section*{Appendix}
\input{sections/appendix}

\end{document}

%% file: sections/abstract.tex
Activation steering turns localized representations into control directions,
but localization alone does not reveal whether a direction has a selective
operating regime. We introduce \emph{Predictive Memory Localization} (PML),
which treats the measured-grid intervention path as the predictive object of
memory localization. PML separates random-calibrated target movement from
semantic-neighbor and capability damage, and compares static localization and
supervised geometry with a strength-disjoint low-dose causal response. Our
frozen study covers 3,000 records from nine datasets and fourteen domains,
yielding 30,000 distinct record--direction--layer paths and 210,000 distinct
path--strength evaluations. At layer 7, the geometry-derived RFM/AGOP direction reaches 13.1\% target-any and
12.3\% clean-any, exceeding random by 3.6 and 3.4 percentage points under a
record-paired bootstrap. Across record-, dataset-, and domain-grouped splits,
responses at $|\alpha|=0.1$ are the strongest signal for outcomes at disjoint strengths
$|\alpha|\in\{0.25,0.5\}$. On held-out records, a predictor-driven selector
chooses a coefficient or abstains, improves utility and reduces
semantic-neighbor damage relative to a train-tuned fixed-strength policy, and
avoids most evaluations in a dense scan.
\input{sections/cross_model_abstract_generated}
PML therefore turns memory localization into a falsifiable forecast of
margin-level selective outcomes and a risk-aware intervention decision.

%% file: sections/cross_model_abstract_generated.tex
Across three residual-norm-matched base models, learned directions retain
selective-path gains and low-dose responses yield $0.801$--$0.828$
record-held-out macro AUROC.

%% file: sections/introduction.tex
\section{Introduction}

Language-model internals have been associated with feed-forward memories,
knowledge neurons, and causally important states
\citep{geva-etal-2021-transformer,dai-etal-2022-knowledge,
NEURIPS2022_6f1d43d5}. This literature suggests an operational hypothesis:
once information is localized, the corresponding representation should provide
a useful control point. Activation steering tests that hypothesis without changing model parameters
\citep{turner2023steering,rimsky-etal-2024-steering,zou2023representation,
pmlr-v238-singh24d}. Yet behavior varies across prompts, layers, models, and
coefficients, and stronger interventions can damage specificity or unrelated
capabilities \citep{NEURIPS2023_81b83900,stoehr-etal-2024-activation,
tan2024steering,goyal-iii-2026-steering}. Steering is thus a multidimensional
control problem \citep{wu25a,pres2024reliable,xu-etal-2026-controllable}.

Localization describes a representation, whereas controllability is revealed
across a \emph{measured intervention path}. Target leverage can coexist with
semantic-neighbor or capability damage, and a clean effect at one coefficient
need not persist at another. Predicting calibrated target, damage, and clean
outcomes over a pre-specified grid asks which localized directions admit a
usable measured coefficient. A single endpoint cannot answer this question:
the same direction may show leverage at one measured strength, damage at
another, or no coefficient where the two separate cleanly.

We introduce \emph{Predictive Memory Localization} (PML), which maps each
record, direction, and layer to those random-calibrated outcomes. It organizes
predictive evidence from baseline belief and metadata, static localization, and
supervised representation geometry to a strength-disjoint low-dose causal
response. This design directly tests whether static evidence forecasts later
behavior or whether an inexpensive causal measurement is required. A
collateral-aware policy then selects a coefficient or abstains.
Unlike generation-level concept prediction \citep{fan2026steerable}, PML audits
localized factual and reasoning directions with explicit collateral probes and
strength-disjoint labels.

\begin{figure*}[t]
    \centering
    \begin{tabular}{@{}c@{\qquad}c@{}}
        \small\textbf{(a) Internal representation} &
        \small\textbf{(b) Forecast utility and decide} \\
        \includegraphics[width=0.45\textwidth,height=2.45in,keepaspectratio]{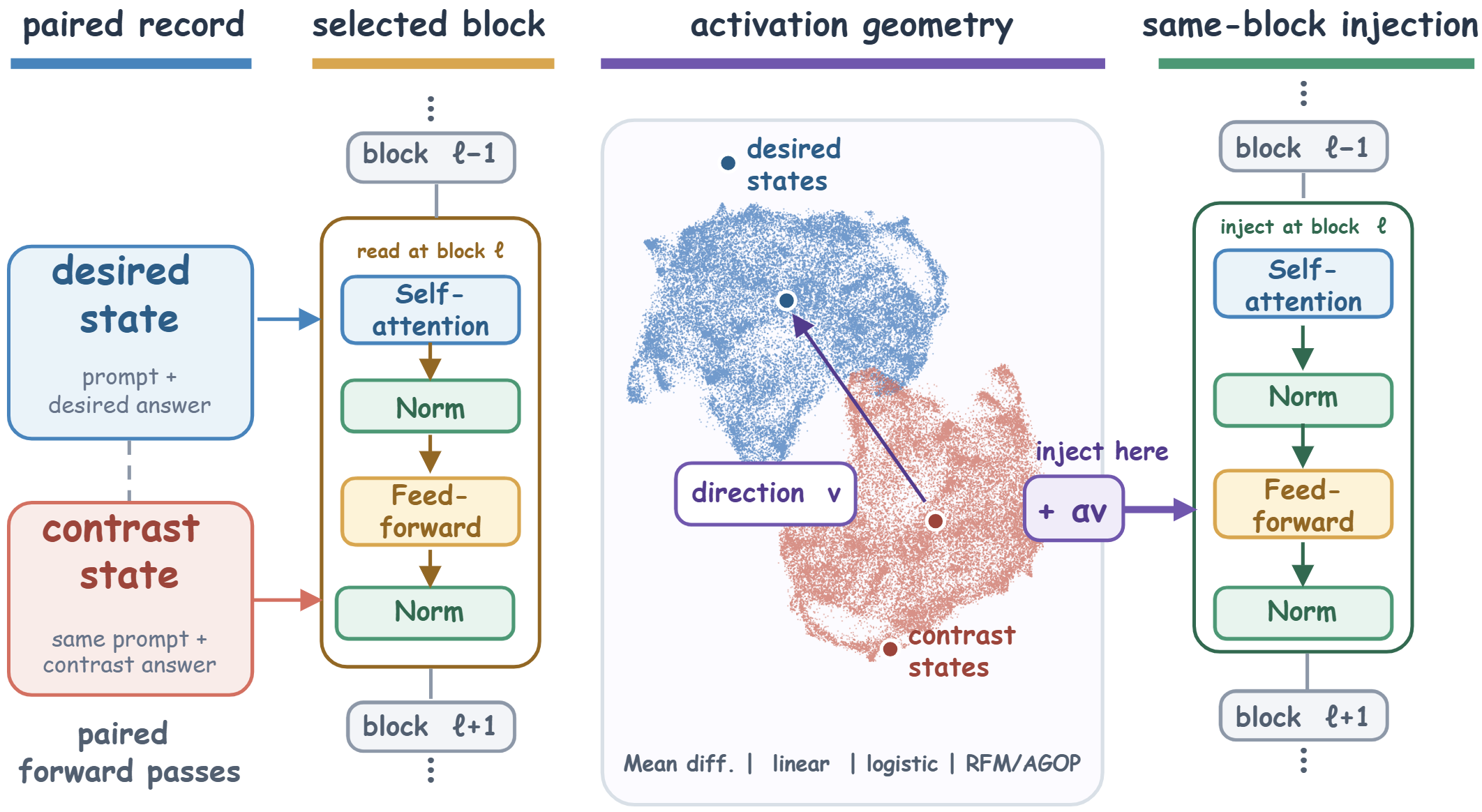} &
        \includegraphics[width=0.45\textwidth,height=2.45in,keepaspectratio]{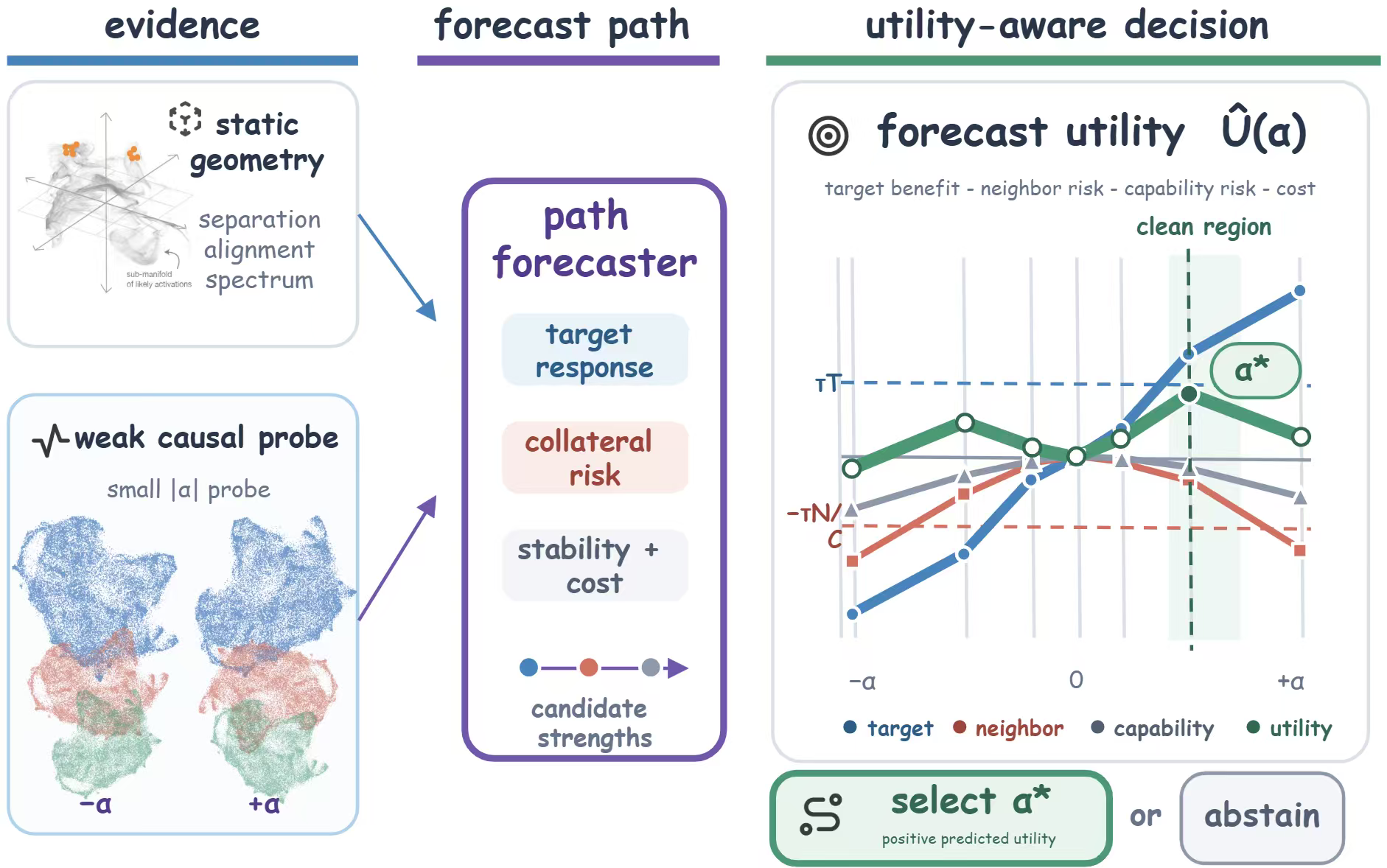}
    \end{tabular}
    \caption{PML from representation to selective control. (a) Direction
    estimators use desired and contrast activations at a selected block, where
    the resulting direction is injected. (b) Static evidence and a disjoint
    low-dose response forecast target benefit, collateral risk, and utility
    over candidate coefficients for selection or abstention. Curves are
    schematic.}
    \label{fig:pml-pipeline}
\end{figure*}

On 3,000 frozen records from nine datasets and fourteen domains, learned
directions improve target and clean-path incidence over random. Static
localization adds little predictive value, whereas a disjoint low-dose response
dominates later-path prediction across grouped transfer. Both findings replicate
across three residual-norm-matched base models (0.801--0.828 record-held-out
macro AUROC). A held-out selector improves utility and reduces neighbor damage
relative to a fixed intervention. The central result is that \textbf{a small
causal response is the most useful forecast of later selective behavior.}

This work makes three contributions:
\begin{itemize}
    \item We define a random-calibrated measured-grid path that jointly records
    target leverage, semantic-neighbor damage, capability damage, and clean
    measured coefficients for each record, direction, and layer.
    \item We separate static localization and supervised geometry from a
    strength-disjoint causal probe, showing that the former are weak predictive
    priors while the latter supplies the dominant signal across grouped and
    cross-model evaluation.
    \item We connect forecasting to action with a held-out collateral-aware
    policy that selects one coefficient or abstains, providing a proof of
    concept for reducing dense intervention evaluation.
\end{itemize}

%% file: sections/related_work.tex
\section{Related Work}

\paragraph{Knowledge representation, localization, and editing.}
Transformer internals have been associated with feed-forward memories,
knowledge neurons, and causally important states
\citep{geva-etal-2021-transformer,dai-etal-2022-knowledge,
NEURIPS2022_6f1d43d5}. Editing uses learned editors, external memory, or
targeted parameter updates
\citep{de-cao-etal-2021-editing,mitchell2022fast,mitchell22a,meng2023mass}, with
efficacy, generalization, and locality organized by surveys and toolkits
\citep{yao-etal-2023-editing,wang-etal-2024-easyedit}.
Localization need not identify components that support editing or unlearning
\citep{NEURIPS2023_3927bbdc,lee-etal-2025-localization}; PML tests its predictive
connection to intervention outcomes.

\paragraph{Activation steering and control tradeoffs.}
Activation engineering intervenes on intermediate representations
\citep{turner2023steering}; contrastive activation addition, representation
engineering, and representation surgery construct or analyze steering
directions from examples and population structure
\citep{rimsky-etal-2024-steering,zou2023representation,pmlr-v238-singh24d}.
Other work studies truthfulness components, coefficient scaling, and
preference--utility tradeoffs
\citep{NEURIPS2023_81b83900,stoehr-etal-2024-activation,
xu-etal-2026-steering}. AxBench compares concept detection and steering methods
\citep{wu25a}, while broader evaluations expose sensitivity to instruction
phrasing, prompt distribution, layer choice, and model scale
\citep{tan2024steering,silva-etal-2025-steering,
goyal-iii-2026-steering}. These studies motivate joint leverage and side-effect
measurement; PML additionally asks whether localization evidence forecasts
their coexistence for each record--direction--layer path.

\paragraph{Evaluation and prediction of steerability.}
Reliable steering evaluations increasingly separate behavioral success from
coherence, specificity, and unintended change
\citep{pres2024reliable,xu-etal-2026-controllable}. Most closely related,
\citet{fan2026steerable} predict under-, successful, or over-steering from
first-token dynamics and rank strengths to reduce rollouts. PML instead predicts
a record--direction--layer path on a pre-specified coefficient grid, calibrates
target, semantic-neighbor, and capability events against random directions, and
optimizes collateral-aware margin utility. Its weak feature is measured at a
coefficient excluded from the stronger-coefficient labels, after which a policy
chooses one coefficient or abstains. SteerBoost targets efficient generation-
level concept alignment; PML tests whether localization and a fixed low-dose
diagnostic forecast selective margin outcomes.

\paragraph{Supervised geometry as a diagnostic.}
Recursive feature machines use the average gradient outer product to learn
task-adapted geometry \citep{radhakrishnan2022mechanism}. PML uses its leading
direction, spectral concentration, and alignment as candidate signals, while
explicitly testing rather than assuming that supervised sensitivity implies
selective control.

%% file: sections/problem_setup.tex
\section{Predictive Memory Localization}

PML connects a localized residual representation to a measured strength-indexed
intervention path and a held-out decision (Figure~\ref{fig:pml-pipeline}). Here,
``memory localization'' denotes a record-conditioned internal signal associated
with an answer relation, not a claim that the relation resides in one unique
physical component.

\subsection{Records, Probes, and Interventions}

Record $i$ contains target prompts $\mathcal{T}_i$, semantic-neighbor prompts
$\mathcal{N}_i$, and general-capability prompts $\mathcal{C}_i$. The first set
expresses the behavior to suppress or enhance; the latter two test whether
nearby knowledge or unrelated abilities are preserved. For prompt $x$ with
desired answer $y^+$ and contrast $y^-$, the answer margin is
\begin{equation}
    m(x)=\log p(y^+\mid x)-\log p(y^-\mid x).
\end{equation}
All outcomes are changes from the unperturbed margin, making paths comparable
across records with different baseline confidence.

At layer $\ell$, method $s$ constructs a unit-norm record-specific direction
$\mathbf{v}_{i\ell s}$. Intervention strength $\alpha$ modifies the residual
activation as
\begin{equation}
    \mathbf{h}'_{\ell}=\mathbf{h}_{\ell}+\alpha\mathbf{v}_{i\ell s}.
\end{equation}
Negative and positive coefficients test suppression and enhancement with the
same direction. We write $\Delta^T_i(\alpha)$ for signed target-margin movement
and $\Delta^N_i(\alpha),\Delta^C_i(\alpha)$ for neighbor and capability
movement; a sufficiently negative collateral change is damage.

\subsection{Random-Calibrated Intervention Paths}

Target, neighbor, and capability responses have different null scales, so
random directions define separate 95th-percentile thresholds $\tau_T$,
$\tau_N$, and $\tau_C$. A target effect crosses $\tau_T$ in the intended
direction, while neighbor or capability damage crosses the corresponding
negative threshold. A \emph{clean strength} produces a target effect while
crossing neither damage threshold at that same coefficient.

Across an ordered, finite strength set, these events form a measured-grid
intervention path. Target and collateral onset identify the first observed
crossing, and adjacent clean strengths form an observed clean region. These are
descriptive grid statistics, not estimates of an unmeasured continuous window.
The confirmatory prediction task focuses on four directly measured events at
held-out stronger coefficients:
\emph{Target-any}, semantic-neighbor damage,
capability damage, and \emph{Clean-any}. Percentages count
record--method--layer paths rather than prompts or individual strengths.

\subsection{Prediction Task}

PML forecasts later path outcomes from progressively stronger evidence.
Baseline belief $B$ and metadata $M$ are augmented with static localization
features $L$ and, for RFM, geometry features $G$. Low-dose response features
$R$ measure target and collateral movement at $\alpha=\pm0.1$. Because $R$
requires intervention, it is a cheap dynamic diagnostic rather than static
localization. The central comparisons test whether $L$ or $G$ improves on
$B+M$, and whether the strength-disjoint response $R$ forecasts outcomes at
$|\alpha|\in\{0.25,0.5\}$. A downstream policy then uses these forecasts to
select a coefficient or abstain.

%% file: sections/method.tex
\section{Signals and Predictive Models}

PML evaluates a common path-prediction and decision pipeline over several
direction families, separating the quality of a direction from the evidence
used to forecast its later behavior.

\subsection{Direction Families}

\paragraph{Random control.}
A seeded unit-norm Gaussian direction defines response thresholds and the
paired null baseline. The logged \texttt{matched\_norm\_random} entry is
numerically identical and retained only for auditability.

\paragraph{Mean difference.}
For positive and contrast activation sets $\mathcal{H}^+$ and
$\mathcal{H}^-$, we use
\begin{equation}
    \mathbf{v}_{\mathrm{mean}}=
    \frac{\boldsymbol{\mu}^+-\boldsymbol{\mu}^-}
    {\|\boldsymbol{\mu}^+-\boldsymbol{\mu}^-\|_2},
\end{equation}
the multi-example analogue of activation addition
\citep{turner2023steering,rimsky-etal-2024-steering}.

\paragraph{Linear and logistic probes.}
Normalized classifier weights provide supervised discriminative directions
without nonlinear feature learning.

\paragraph{RFM/AGOP direction.}
A recursive feature machine estimates task-adapted geometry through the average
gradient outer product \citep{radhakrishnan2022mechanism}. Its leading
eigenvector provides a low-rank direction, while spectrum, concentration, and
alignment statistics become geometry features. A top-$k$ boundary study tests
whether broader supervised subspaces trade selectivity for leverage.

All directions are fitted independently per record and layer and injected at
the selected residual block during evaluation. Detailed activation extraction,
fitting hyperparameters, inference settings, and continuation scoring are
included in the released protocol.

\subsection{Predictive Evidence and Grouped Evaluation}

Static localization $L$ summarizes class separation, projection, saliency, and
direction agreement; RFM geometry $G$ summarizes spectral concentration and
alignment. These features are nested with baseline belief $B$, metadata $M$,
and the low-dose response $R$ so that method identity or an observed response
cannot be misattributed to static localization.

We fit class-balanced logistic regression and random forests for target,
damage, and clean outcomes. Five-fold evaluation groups all paths from the same
record, dataset, or domain, preventing related trajectories from crossing a
train--test boundary. We report prevalence, AUROC, and average precision in the
main paper;
additional classification and calibration metrics are in the released
artifacts.

\subsection{Risk-Aware Strength Selection}

For each nonzero candidate coefficient, the selector predicts clean-effect
probability and continuous utility. With $q(\alpha)\in\{-1,+1\}$ denoting the
requested suppression or enhancement sign, measured utility is
\begin{equation}
\begin{aligned}
    u_i(\alpha)={}&q(\alpha)\Delta^T_i(\alpha)
    -\max\{0,-\Delta^N_i(\alpha)\}\\
    &-\max\{0,-\Delta^C_i(\alpha)\}.
\end{aligned}
\end{equation}
Target movement is rewarded and collateral margin decreases receive unit
penalties. The decision score is
\begin{equation}
    \widehat{u}_i(\alpha)+0.1\widehat{p}_i(\mathrm{clean}\mid\alpha)
    -0.01|\alpha|.
\end{equation}
The policy selects the highest-scoring coefficient and abstains when its score
is nonpositive. This utility is defined on teacher-forced answer-margin changes,
not free-generation correctness. We evaluate it against both no intervention,
whose utility is zero by construction, and a train-tuned fixed coefficient.

%% file: sections/experiments.tex
\section{Experiments}

We organize the evidence around three questions: whether learned directions
create selective intervention paths, which internal signals forecast those
paths, and whether those forecasts improve strength decisions. We report the
frozen confirmatory study here; the Supplementary Material adds protocol
details, uncertainty estimates, and analyses not shown in the main paper.

\subsection{Frozen Multidomain Study}

The benchmark contains 3,000 records from nine public sources: MMLU-Pro,
MMLU-Redux 2.0, AI2 ARC, OpenBookQA, SciQ, LiveBench reasoning and math,
HellaSwag, and QASC
\citep{wang2024mmlupro,gema2024mmluredux,clark2018arc,
mihaylov-etal-2018-suit,welbl-etal-2017-crowdsourcing,white2025livebench,
zellers-etal-2019-hellaswag,khot2020qasc}. The collection spans fourteen
academic, scientific, commonsense, mathematical, and reasoning domains. A
schema-constrained generation step converts each source item into disjoint
direction-fitting statements, three record-specific target probes, and three
semantic-neighbor probes; four globally balanced capability probes are assigned
per record. The worked example below traces one source item from fitting
evidence to evaluation probes and its resulting path label. Dataset composition,
validation, and additional examples appear in the Supplementary Material.

\input{tables/main_worked_record}

We evaluate Qwen3-1.7B-Base \citep{yang2025qwen3} at two middle-depth
Transformer blocks, reported as blocks 7 and 11 by the model implementation and
selected using a 500-record layer-selection subset
(Table~\ref{tab:layer-selection}). We
compare five direction constructions over a signed coefficient sweep. Random
directions calibrate target and collateral thresholds, while the weak response
at $|\alpha|=0.1$ is disjoint from the stronger coefficients used to define
confirmatory outcomes. Table~\ref{tab:main-results} defines the reported path
rates, and prediction folds hold out complete records, datasets, or domains.

For cross-model confirmation, the same frozen 500-record subset is evaluated
on Qwen3-1.7B, Qwen3.5-2B-Base \citep{qwen2026qwen35}, and
Ministral-3-3B-Base \citep{liu2026ministral3}. We align relative
intervention budget $\|\alpha v\|_2/\|h\|_2$ using
\begin{equation}
    s_{m,\ell}=\frac{\operatorname{RMS}(h_{m,\ell})}
    {\operatorname{RMS}(h_{\mathrm{ref},\ell})}
    \sqrt{\frac{d_m}{d_{\mathrm{ref}}}},
    \qquad \alpha_{m,\ell}=s_{m,\ell}\alpha_{\mathrm{ref}}.
    \label{eq:rms-matching}
\end{equation}
Scales are fixed on a disjoint 100-record calibration set, after which each
model recalibrates its random-response thresholds on the formal cohort. The
confirmation retains random, mean-difference, logistic, and RFM/AGOP directions
at two pre-specified blocks per model. We omit linear because it is not strongest
at either primary-study block, which preserves a symmetric comparison across
architectures.

% Queue the single-column layer diagnostic before the full-width outcome table
% while preserving the manuscript's Table 1/2 numbering.
\setcounter{table}{1}
\input{tables/layer_selection}
\setcounter{table}{0}
\input{tables/main_results}
\setcounter{table}{2}

\subsection{Learned Directions Improve Selective Paths}

Table~\ref{tab:main-results} reports both the powered 3,000-record outcome
estimate and the three pre-specified 500-record confirmations. Its entries are
path-incidence percentages; $\Delta$T and $\Delta$C are percentage-point
differences from the within-model random control, not AUROC. In the primary
study, layer-7 RFM/AGOP reaches \textbf{13.1\% Target and 12.3\% Clean}, the
strongest selective-path result. More broadly, learned directions improve
target leverage and clean-path incidence over random controls, although the
best construction depends on the block.

Record-paired bootstrap intervals confirm the learned-over-random Target and
Clean gains for the strongest primary-study directions. Collateral-damage
intervals include zero, so the evidence supports more usable intervention paths
rather than a universal reduction in every form of collateral movement. Full
intervals are reported in the supplement.

\begin{figure*}[!t]
    \centering
    \includegraphics[width=0.89\textwidth]{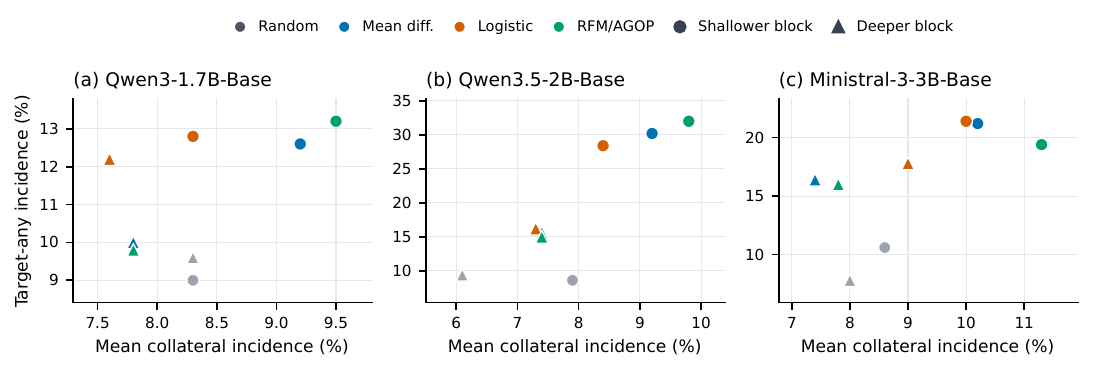}
    \caption{Target leverage and collateral incidence on the common 500-record
    cohort. Each panel shows one base model; colors identify direction
    construction and marker shapes identify the shallower or deeper
    pre-specified block. The vertical axis is Target-any path incidence, while
    the horizontal axis averages semantic-neighbor and capability-damage
    incidence. Points are not connected because the two blocks are separately
    pre-specified evaluations rather than a continuous trajectory. Axes are
    panel-specific so that within-model trade-offs remain visible.}
    \label{fig:path-outcome-profiles}
\end{figure*}

The residual-norm-matched confirmations preserve the same qualitative pattern
at model-dependent magnitudes. Figure~\ref{fig:path-outcome-profiles} shows that
learned directions generally move upward from their random controls in target
leverage, but not uniformly leftward toward lower collateral incidence. The
Qwen3-1.7B subset also closely tracks the powered estimate, separating cohort
variation from the cross-model scale alignment.

\subsection{Low-Dose Responses Forecast Later Outcomes}

\begin{figure*}[!t]
    \centering
    \includegraphics[width=0.97\textwidth]{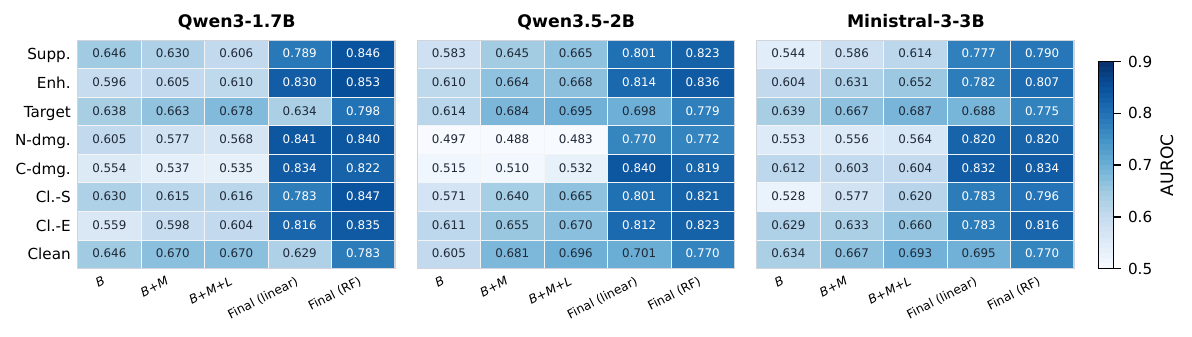}
    \caption{Cross-model path prediction. Per-outcome record-held-out
    AUROC on the common 500-record cohort. The first three random-forest columns
    cumulatively add base margins and record descriptors ($B$), method/block
    metadata ($M$), and static localization ($L$). The final two columns use the
    complete $B{+}M{+}L{+}R$ features with a linear predictor or random forest,
    providing a predictor-class ablation. The weak response produces the
    dominant feature gain under both predictors, while the nonlinear model gives
    the strongest final performance.}
    \label{fig:cross-model-prediction-heatmap}
\end{figure*}

Prediction is the central test of PML. Static localization is a weak prior:
Figure~\ref{fig:cross-model-prediction-heatmap} establishes a consistent
diagnostic hierarchy. Static localization $L$ adds little beyond base and metadata
features, and supervised geometry $G$ does not change that conclusion. In
contrast, the strength-disjoint weak response $R$ produces the dominant gain
across outcomes and models. Table~\ref{tab:cpu-controls} further separates this
gain from ordinary scalar weak-to-strong correlation. Positive-label prevalence
is only 8--11\%, so the table reports AP with AUROC. For Target-any and
Clean-any, a multivariate $R$-only random forest substantially improves on a
single signed response, with the complete predictor adding a smaller final
gain. Final macro AUROC remains around \textbf{0.80--0.85} under record-,
dataset-, and domain-held-out evaluation. Removing all 500 records used for
layer selection leaves the four principal full-predictor AUROCs within 0.01 of
the 3,000-record estimates. Thus, observing a structured low-dose causal
response is substantially more informative about the later intervention path
than static localization geometry alone, and this conclusion is not explained
by the layer-selection overlap.

\FloatBarrier

\subsection{Forecasts Improve Strength Decisions}

We train candidate-strength models from complementary dense and sparse
trajectories, then evaluate on disjoint dense-grid records. Before a candidate
outcome is revealed, each policy selects one coefficient or abstains.
Table~\ref{tab:main-strength-selector} gives a 100-record proof of concept.
Relative to a train-tuned fixed coefficient, utility improves by \textbf{0.055}
and \textbf{0.034}, mainly as neighbor damage falls from 6.0\% to 1.8\% and
5.2\% to 2.2\%. Suppression exceeds no intervention; enhancement does not
significantly do so. The policy averages 2.55/2.61 coefficient evaluations---two
weak probes plus a final action---versus 26 for a dense scan; shared direction
fitting is excluded. Weight sensitivity and full outcomes are supplementary.

\subsection{Endpoint Scope}

Free-generation stress tests are reported only in the Supplementary Material.
They show that margin movement can alter text but does not yield reliable
wrong-to-right correction; PML's primary evidence is therefore margin-level,
not a claim of stable generated-answer control.

%% file: tables/main_worked_record.tex
\begin{center}
\centering
\small
\setlength{\tabcolsep}{2.4pt}
\begin{tabular}{@{}p{0.20\columnwidth}p{0.74\columnwidth}@{}}
\toprule
Object & Frozen example \\
\midrule
Source & MMLU-Redux 2.0 chemistry: Suppose that the 13C nuclei in a molecule in a 600 MHz spectrometer can be 100\% polarized (p = 1). If T1 = 5.0 s, how long does it take for p to reach a value equal to twice the thermal equilibrium polarization at 298 K? \\
Fit evidence & Positive statement: The polarization reaches twice the thermal equilibrium value in 72.0 seconds. Contrast statement: The polarization reaches twice the thermal equilibrium value in 56.6 seconds. \\
Target probe & Prompt: After full polarization, how long until it reaches twice thermal equilibrium?; candidate answers: 72.0 s vs. 56.6 s. \\
Neighbor probe & Prompt: The spin-lattice relaxation time T1 for 13C in this experiment is; candidate answers: 5.0 s vs. 10.0 s. \\
Capability probe & Prompt: An object in motion tends to stay in motion unless acted upon by an external force. This is; candidate answers: Newton's first law vs. Newton's second law. \\
Measured path & Mean difference, layer 11: suppression onset $-0.25$; positive neighbor onset $0.5$; no enhancement or capability onset. \\
Labels & Target=1, N-dmg.=1, C-dmg.=0, Clean suppression=1, Clean=1. \\
\bottomrule
\end{tabular}
\captionof*{table}{\textbf{Worked example.} One frozen record from construction to path label; direction-fitting statements and evaluation probes are disjoint.}
\end{center}

%% file: tables/layer_selection.tex
\begin{table}[!t]
\centering
\includegraphics[width=0.96\columnwidth]{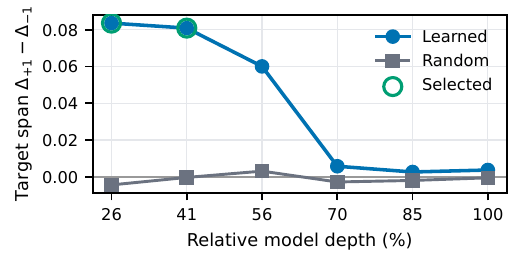}
\caption{Layer selection on a 500-record subset. Target-margin span across relative depth identifies blocks 7 and 11 as the strongest distinct middle-depth blocks; random spans remain near zero.}
\label{tab:layer-selection}
\end{table}

%% file: tables/main_results.tex
\begin{table*}[!t]
\centering
\small
\renewcommand{\arraystretch}{0.88}
\setlength{\tabcolsep}{1.4pt}
\begin{tabular*}{\textwidth}{@{\extracolsep{\fill}}clrrrrrrrrrr@{}}
\toprule
Layer & Direction & Supp. & Enh. & Target & N-dmg. & C-dmg. & Cl. supp. & Cl. enh. & Clean & $\Delta$T & $\Delta$C \\
\midrule
\multicolumn{12}{c}{\textbf{A. Qwen3-1.7B-Base} \quad ($n=3,000$, primary scale, $\tau_T=0.175$)} \\
\cmidrule(lr){1-12}
7 & Random & 5.1 & 4.7 & 9.5 & 8.2 & 8.2 & 4.7 & 4.3 & 8.9 & -- & -- \\
7 & Mean diff. & 6.1 & 7.0 & 12.8 & 7.9 & 8.2 & 5.6 & 6.7 & 12.0 & +3.3 & +3.2 \\
7 & Logistic & 6.7 & 6.1 & 12.4 & 9.1 & 9.0 & 6.3 & 5.7 & 11.6 & +2.8 & +2.8 \\
7 & Linear & 5.8 & 6.2 & 11.6 & 8.8 & 8.4 & 5.2 & 6.0 & 10.8 & +2.1 & +2.0 \\
7 & RFM/AGOP & 7.1 & 6.4 & \textbf{13.1} & 8.5 & 9.0 & 6.5 & 6.1 & \textbf{12.3} & \textbf{+3.6} & \textbf{+3.4} \\
\addlinespace[0.35pt]
11 & Random & 4.7 & 4.2 & 8.8 & 7.0 & 7.1 & 4.5 & 3.8 & 8.3 & -- & -- \\
11 & Mean diff. & 6.2 & 5.3 & \textbf{11.3} & 7.8 & 7.1 & 5.6 & 5.1 & \textbf{10.6} & \textbf{+2.5} & \textbf{+2.3} \\
11 & Logistic & 5.7 & 5.7 & 11.2 & 7.4 & 7.1 & 5.4 & 5.4 & \textbf{10.6} & +2.4 & \textbf{+2.3} \\
11 & Linear & 5.3 & 5.1 & 10.2 & 6.8 & 7.1 & 5.0 & 4.9 & 9.7 & +1.4 & +1.4 \\
11 & RFM/AGOP & 5.1 & 5.3 & 10.2 & 7.5 & 7.2 & 4.8 & 5.1 & 9.7 & +1.4 & +1.5 \\
\midrule
\multicolumn{12}{c}{\textbf{B. Qwen3-1.7B-Base} \quad ($n=500$, residual-norm matched, $\tau_T=0.177$)} \\
\cmidrule(lr){1-12}
7 & Random & 4.4 & 4.8 & 9.0 & 7.4 & 9.2 & 4.0 & 4.0 & 7.8 & -- & -- \\
7 & Mean diff. & 5.2 & 7.4 & 12.6 & 8.2 & 10.2 & 4.4 & 7.2 & 11.6 & +3.6 & +3.8 \\
7 & Logistic & 6.8 & 6.2 & 12.8 & 8.0 & 8.6 & 6.6 & 5.6 & 12.0 & +3.8 & +4.2 \\
7 & RFM/AGOP & 6.2 & 7.8 & \textbf{13.2} & 9.4 & 9.6 & 5.6 & 7.2 & \textbf{12.4} & \textbf{+4.2} & \textbf{+4.6} \\
\addlinespace[0.35pt]
11 & Random & 4.8 & 4.8 & 9.6 & 7.6 & 9.0 & 4.8 & 4.4 & 9.2 & -- & -- \\
11 & Mean diff. & 6.0 & 4.6 & 10.0 & 7.6 & 8.0 & 5.0 & 4.2 & 8.6 & +0.4 & -0.6 \\
11 & Logistic & 5.8 & 6.4 & \textbf{12.2} & 7.8 & 7.4 & 5.6 & 5.8 & \textbf{11.4} & \textbf{+2.6} & \textbf{+2.2} \\
11 & RFM/AGOP & 4.2 & 5.8 & 9.8 & 8.2 & 7.4 & 3.8 & 5.2 & 8.8 & +0.2 & -0.4 \\
\midrule
\multicolumn{12}{c}{\textbf{C. Qwen3.5-2B-Base} \quad ($n=500$, residual-norm matched, $\tau_T=0.111$)} \\
\cmidrule(lr){1-12}
6 & Random & 4.6 & 4.0 & 8.6 & 7.6 & 8.2 & 3.6 & 3.6 & 7.2 & -- & -- \\
6 & Mean diff. & 16.6 & 17.2 & 30.2 & 9.6 & 8.8 & 14.6 & 16.0 & 28.2 & +21.6 & +21.0 \\
6 & Logistic & 18.8 & 16.0 & 28.4 & 11.2 & 5.6 & 17.0 & 15.2 & 26.8 & +19.8 & +19.6 \\
6 & RFM/AGOP & 18.8 & 17.6 & \textbf{32.0} & 10.6 & 9.0 & 17.8 & 17.4 & \textbf{31.2} & \textbf{+23.4} & \textbf{+24.0} \\
\addlinespace[0.35pt]
9 & Random & 4.6 & 4.8 & 9.4 & 6.8 & 5.4 & 4.2 & 4.4 & 8.6 & -- & -- \\
9 & Mean diff. & 9.2 & 7.2 & 15.6 & 9.6 & 5.2 & 8.8 & 6.8 & \textbf{15.0} & +6.2 & \textbf{+6.4} \\
9 & Logistic & 7.8 & 8.8 & \textbf{16.2} & 8.6 & 6.0 & 7.0 & 8.0 & 14.6 & \textbf{+6.8} & +6.0 \\
9 & RFM/AGOP & 8.0 & 7.4 & 15.0 & 9.0 & 5.8 & 7.4 & 6.6 & 13.8 & +5.6 & +5.2 \\
\midrule
\multicolumn{12}{c}{\textbf{D. Ministral-3-3B-Base} \quad ($n=500$, residual-norm matched, $\tau_T=0.083$)} \\
\cmidrule(lr){1-12}
6 & Random & 6.2 & 5.0 & 10.6 & 7.8 & 9.4 & 5.6 & 4.6 & 9.8 & -- & -- \\
6 & Mean diff. & 11.2 & 11.2 & 21.2 & 10.6 & 9.8 & 10.4 & 10.6 & 19.8 & +10.6 & +10.0 \\
6 & Logistic & 10.0 & 12.2 & \textbf{21.4} & 11.2 & 8.8 & 9.8 & 11.6 & \textbf{20.6} & \textbf{+10.8} & \textbf{+10.8} \\
6 & RFM/AGOP & 11.0 & 9.8 & 19.4 & 11.2 & 11.4 & 10.4 & 9.2 & 18.2 & +8.8 & +8.4 \\
\addlinespace[0.35pt]
10 & Random & 3.6 & 4.6 & 7.8 & 7.2 & 8.8 & 3.2 & 4.6 & 7.4 & -- & -- \\
10 & Mean diff. & 9.0 & 8.0 & 16.4 & 8.4 & 6.4 & 8.8 & 7.4 & 15.6 & +8.6 & +8.2 \\
10 & Logistic & 9.6 & 8.6 & \textbf{17.8} & 8.8 & 9.2 & 9.0 & 8.2 & \textbf{16.8} & \textbf{+10.0} & \textbf{+9.4} \\
10 & RFM/AGOP & 8.6 & 8.2 & 16.0 & 7.6 & 8.0 & 8.0 & 7.4 & 14.8 & +8.2 & +7.4 \\
\bottomrule
\end{tabular*}
\caption{Random-calibrated path incidence (\%) for the 3,000-record primary study and three 500-record confirmations. N/C-dmg. are collateral damage; $\Delta$ columns are learned-minus-random points. Bold marks each block's strongest learned result.}
\label{tab:main-results}
\end{table*}

%% file: sections/discussion.tex
\section{Discussion}

\paragraph{Measured paths separate leverage from selectivity.}
Learned directions increase Target and Clean incidence, yet their neighbor- and
capability-damage differences remain statistically unresolved. A direction can
therefore create more usable measured coefficients without becoming uniformly
safer. Layer 7 similarly offers greater leverage together with more collateral
movement than layer 11. The relevant object is not maximal sensitivity but the
coexistence of target and damage responses on the same pre-specified grid.
Measured clean regions identify where leverage and selectivity coincide, but
they should not be read as broad continuous operating windows: most observed
clean paths contain only one measured clean coefficient, and strict
target-first or damage-first orderings are rare. These topology statistics are
therefore diagnostics that motivate multi-strength evaluation rather than the
primary prediction labels.

\input{tables/main_cpu_controls}
\input{tables/main_strength_selector}

\input{sections/cross_model_discussion_generated}

\paragraph{A small causal response is more actionable than static localization.}
Supervised geometry is valuable for constructing high-leverage directions and
describing their concentration and alignment, but these static descriptors add
little forecasting power by themselves. In contrast, a strength-disjoint
low-dose response strongly predicts later outcomes across record, dataset, and
domain transfer. This suggests that localization becomes actionable when it is
paired with a cheap causal measurement of the specific path, rather than when
static separation is treated as sufficient evidence of control. PML therefore
complements concept detection and average steering evaluation
\citep{wu25a,silva-etal-2025-steering,goyal-iii-2026-steering} by testing
whether an internal signal supports effective and selective margin movement at
a chosen strength.

The weak, outcome-specific transfer of activation-path features to ROME marks a
second boundary. A path can diagnose a fragile or promising activation
intervention without identifying the best persistent parameter edit
\citep{NEURIPS2023_3927bbdc}; PML does not equate activation controllability
with editability.

%% file: tables/main_cpu_controls.tex
\begin{table}[t]
\centering
\small
\renewcommand{\arraystretch}{1.05}
\setlength{\tabcolsep}{2.5pt}
\begin{tabular*}{\columnwidth}{@{\extracolsep{\fill}}lrrrr@{}}
\toprule
Outcome & Prev. & Scalar & $R$-only RF & Full RF \\
\midrule
Target-any & .111 & .708/.324 & .793/\textbf{.367} & \textbf{.815}/.357 \\
Clean-any & .105 & .651/.266 & .789/\textbf{.342} & \textbf{.810}/.330 \\
Neighbor dmg. & .079 & .840/\textbf{.408} & .842/.402 & \textbf{.858}/.396 \\
Capability dmg. & .078 & .849/.373 & .842/.373 & \textbf{.856/.382} \\
\bottomrule
\end{tabular*}
\caption{Record-held-out prediction on Qwen3-1.7B-Base. Cells are AUROC/AP;
Prev. is label prevalence. Scalar is training-free, $R$-only uses the
multivariate weak profile, and Full uses $B+M+L+R$.}
\label{tab:cpu-controls}
\end{table}

%% file: tables/main_strength_selector.tex
\begin{table}[t]
\centering
\small
\renewcommand{\arraystretch}{1.05}
\setlength{\tabcolsep}{1.5pt}
\newcommand{\selectorcell}[1]{\raisebox{4pt}{#1}}
\begin{tabular*}{\columnwidth}{@{\extracolsep{\fill}}lcccc@{}}
\toprule
\selectorcell{Objective} & \shortstack{$\Delta U$\\vs. 0} & \shortstack{$\Delta U$\\vs. fixed} & \selectorcell{N-dmg.} & \selectorcell{Abstain} \\
\midrule
\selectorcell{Suppression} & \shortstack{\textbf{.013}\\{[.003,.025]}} & \shortstack{\textbf{.055}\\{[.044,.066]}} & \selectorcell{.060$\to$\textbf{.018}} & \selectorcell{.454} \\
\selectorcell{Enhancement} & \shortstack{.008\\{[$-.001$,.019]}} & \shortstack{\textbf{.034}\\{[.024,.044]}} & \selectorcell{.052$\to$\textbf{.022}} & \selectorcell{.391} \\
\bottomrule
\end{tabular*}
\caption{Held-out decisions on 100 records and 900 paths per objective. Utility
differences use a paired record bootstrap; N-dmg. is fixed$\to$selector.}
\label{tab:main-strength-selector}
\end{table}

%% file: sections/cross_model_discussion_generated.tex
\paragraph{Budget matching preserves the diagnostic hierarchy.}
Residual-norm matching aligns relative intervention magnitude, not response
rates: Qwen3.5 shows the largest gains and Ministral is intermediate. All three
models nevertheless reproduce more learned clean paths and a much larger
predictive contribution from low-dose response than static localization.

%% file: sections/limitations.tex
\section{Limitations}

The primary study uses Qwen3-1.7B and two blocks selected on a 500-record
subset. Width-corrected residual-norm confirmations add Qwen3.5-2B and
Ministral-3-3B at two pre-specified aligned blocks, but they support claims
about those relative depths rather than global layer optimality. Because every
model recalibrates its own random null, cross-model evidence establishes
within-model contrasts and predictor ordering, not absolute incidence
comparisons across architectures. The Qwen3-1.7B subset differs slightly from
the 3,000-record estimate because of sampling and threshold recalibration.

The primary outcomes are teacher-forced margins at two held-out stronger
coefficients. Finite-grid topology and free-generation stress tests are
supplementary; the latter do not establish reliable wrong-to-right control.
Schema-constrained probes receive an independent expert audit and exclusion
sensitivity, but independently authored validation remains future work. The
paired differences in neighbor and capability damage remain unresolved, and
Static localization and AGOP geometry also provide limited incremental
prediction once a weak response is observed. Their present value is therefore
structural---direction construction, concentration, and alignment
diagnostics---rather than a standalone guarantee of path quality.

The response thresholds are frozen global percentiles from random-direction
controls. This supplies a common within-study null, but it does not establish
that the same numeric thresholds transport to a new model or domain. Likewise,
the strength policy optimizes one declared utility with unit collateral
penalties, a 0.1 clean bonus, and a 0.01 magnitude penalty. Weight sensitivity
preserves gains over the fixed policy, not universally over no intervention.

Finally, the 100-record selector is a proof of concept. Its cost reduction
counts coefficient evaluations---two weak probes and a selected action---while
excluding shared direction fitting and feature extraction. Neighbor and
capability probes operationalize two collateral channels but cannot exhaust
downstream side effects; the small ROME transfer study is also insufficient for
claims about general editing success or model-wide safety.

%% file: sections/conclusion.tex
\section{Conclusion}

Predictive Memory Localization forecasts measured-grid target, neighbor,
capability, and clean margin outcomes. On 3,000 records, learned directions
improve Target and Clean incidence over random; at block 7, RFM/AGOP reaches
13.1\% Target and 12.3\% Clean. A strength-disjoint low-dose response dominates
static localization for later-outcome prediction, and the diagnostic hierarchy
replicates across three matched base models. A held-out selector then improves
utility over a fixed coefficient, reduces neighbor damage, and replaces a dense
scan with two weak probes and a selected action or abstention. PML thus converts
a static localization claim into a falsifiable forecast and a risk-aware
decision while making its margin-level and finite-grid scope explicit.

The broader implication is that representation evidence and intervention
quality are related but distinct. A direction can be well localized without
providing a selective operating regime, whereas a small causal response can
reveal leverage and collateral risk before a stronger action. Random-direction
calibration makes this distinction measurable and exposes cases where
abstention is appropriate.

%% file: sections/appendix.tex
\section{Frozen Benchmark Construction}

The frozen benchmark is constructed from nine public multiple-choice and
reasoning sources (Table~\ref{tab:dataset-composition}). Records retain their
source dataset, domain, release year, and freshness group. The final collection
contains 1,950 records from 2024 sources and 1,050 records from pre-2024
sources. Fourteen domains range from situated commonsense and elementary
science to recent corrected facts, academic questions, mathematics, and
reasoning.

The source benchmarks are MMLU-Pro and MMLU-Redux 2.0
\citep{wang2024mmlupro,gema2024mmluredux}, AI2 ARC, OpenBookQA, and SciQ
\citep{clark2018arc,mihaylov-etal-2018-suit,welbl-etal-2017-crowdsourcing},
LiveBench \citep{white2025livebench}, HellaSwag
\citep{zellers-etal-2019-hellaswag}, and QASC \citep{khot2020qasc}.

Each record contains a direction-fitting set and an evaluation assignment. The
direction-fitting statements never contain the target, neighbor, or capability
evaluation fields. Evaluation uses three target probes, three semantic-neighbor
probes, and four capability probes per record. Capability assignments are
round-robin balanced over 2,834 unique probes; every capability probe is used
four or five times. This prevents a small set of easy generic questions from
dominating the capability-damage estimate.

\subsection{Independent Expert Semantic Audit}

Multiple human experts independently reviewed a dataset-stratified sample of
108 complete records, 12 from each source. They checked the desired and contrast
answers, direction-fitting statements, three target probes, three semantic
neighbors, and four capability probes. Sixty-five records pass without issue,
32 have a minor issue that does not reverse the intended relation, and 11 fail
at least one semantic criterion. This yields an 89.8\% acceptable rate on the
audited sample. Because the sample is diagnostic rather than a population error
estimate, we additionally remove the known failed records and then all known
non-pass records from the full CPU analyses. Table~\ref{tab:semantic-audit-sensitivity}
shows that principal predictor AUROC changes by at most 0.004. At block 7, all
learned Target and Clean gains over random also retain positive paired 95\%
intervals under both exclusions; for RFM/AGOP, the most conservative exclusion
gives $+0.0369$ Target and $+0.0352$ Clean. Thus the reported effects are not
driven by the audited exceptions, while independently authored validation
remains important future work.

\input{tables/appendix_dataset_composition}

\section{Confirmatory Experimental Details}

We evaluate Qwen3-1.7B-Base~\cite{yang2025qwen3} at 0-indexed layers 7 and 11, selected from a dense
pilot over layers 7, 11, 15, 19, 23, and 27. The main grid is
$\{-0.5,-0.25,-0.1,0,0.1,0.25,0.5\}$. Five distinct direction families plus
a matched-random audit entry, which is numerically identical to the random
control, and two layers produce 30,000 distinct record--method--layer paths and
36,000 logged rows. Seven coefficients and ten probes per record produce
210,000 distinct and 252,000 logged path--strength rows, plus 2.52 million
logged probe-level rows. Equivalently, the audit artifacts contain 360,000
seven-point probe trajectories, of which 300,000 are distinct after removing
the duplicate entry.
GPU inference uses one NVIDIA GeForce RTX 5090 with bfloat16, batch size two,
and seed 113.

\paragraph{Direction construction and inference.}
Each direction is fitted independently for one record and layer using 6--13
positive and 6--13 contrast prompts. Prompts are tokenized without added
special tokens and left padded; the direction-fitting activation is the final
non-padding prompt token at the selected layer. Mean difference, ridge linear,
and logistic use the same activation matrix and labels. Ridge regularization is
$10^{-3}$, logistic regression runs for at most 1,000 iterations, and RFM uses
three iterations with a Laplace kernel of bandwidth 10 and regularization
$10^{-3}$. During evaluation, the unit direction is added to the selected
block output at every token position. Correct and contrast continuations are
teacher-forced separately, and path labels use their mean per-token log-probability
margin to reduce continuation-length effects.

Random-direction 95th percentiles define separate response thresholds:
$\tau_T=0.1753$, $\tau_N=0.1567$, and $\tau_C=0.1248$. The confirmatory labels
use only $|\alpha|\in\{0.25,0.5\}$. Weak-response features use only
$|\alpha|=0.1$, so the observed probe strength is disjoint from all label
strengths.

Feature group $B$ contains unperturbed target, neighbor, and capability margins.
Group $M$ contains method, layer, dataset, domain, freshness, and release year.
Group $L$ contains separation, saliency, threshold-accuracy, and
direction-agreement statistics, while $G$ contains RFM/AGOP spectrum and
alignment statistics. Group $R$ contains signed target, neighbor, and
capability responses at $\alpha=\pm0.1$.
Prediction uses class-balanced logistic regression and random forests with five
grouped folds by record, dataset, or domain. We report AUROC, average precision,
balanced accuracy, F1, and Brier score in the released result artifacts; the
tables below retain AUROC and AP, the two threshold-independent ranking metrics.

\paragraph{Computational environment.}
\input{sections/computational_environment_generated}

\input{tables/appendix_semantic_audit_sensitivity}

\section{Dimension-Corrected Cross-Model Protocol}

The confirmation uses one seed-113 stratified subset of 500 frozen records for
Qwen3-1.7B-Base, Qwen3.5-2B-Base~\cite{qwen2026qwen35}, and
Ministral-3-3B-Base~\cite{liu2026ministral3}. A separate seed-127
calibration set contains 100 records and has zero overlap with the formal
subset. At each aligned layer, 1,941 direction-construction prompt states are
used to estimate the median residual RMS. The calibration set is used only to
fix model--layer scales, not to select records, methods, or outcomes.

Directions have unit $\ell_2$ norm. Consequently, matching only the numerical
residual RMS would not align the intervention relative to the residual-state
$\ell_2$ norm when hidden widths differ. We match
$\|\alpha v\|_2/\|h\|_2$ with
\begin{equation}
    s_{m,\ell}=\frac{\operatorname{RMS}(h_{m,\ell})}
    {\operatorname{RMS}(h_{\mathrm{ref},\ell})}
    \sqrt{\frac{d_m}{d_{\mathrm{ref}}}},
    \qquad \alpha_{m,\ell}=s_{m,\ell}\alpha_{\mathrm{ref}}.
    \label{eq:supp-rms-matching}
\end{equation}
Qwen3-1.7B and
Qwen3.5 have hidden width 2,048; Ministral has width 3,072 and therefore
includes a $\sqrt{3{,}072/2{,}048}$ correction. The resulting layerwise scales
are 1.000/1.000 for Qwen3-1.7B layers 7/11, 0.088/0.041 for Qwen3.5 layers
6/9, and 0.091/0.052 for Ministral layers 6/10. The manifest stores the RMS
ratio, hidden-width ratio, scale, record hash, and source-summary hashes.

All models retain the same target, neighbor, and capability probes; reference
coefficient grid; and confirmatory outcome definitions. At both aligned layers
we evaluate random, mean difference, logistic, and RFM/AGOP, for 24
model--layer--method configurations, 12,000 record-level paths, and 84,000
path--strength evaluations. Linear remains in the complete 3,000-record primary
study but is omitted here because logistic represents the same supervised
discriminative direction family. Each model recalibrates target, neighbor, and
capability thresholds from its own random directions on the frozen 500-record
cohort. Qwen3-1.7B has scale one, so its 500-record outcomes are exact subsets
of the completed primary run; only the cohort-level random thresholds are
recomputed.

The corresponding path-incidence outcomes are reported in the main paper. This
section records the calibration and normalization choices needed to reproduce
that comparison without duplicating the main result table.

\section{Uncertainty and Worked Examples}

For Table~\ref{tab:paired-bootstrap}, each learned row is paired with the random
row for the same record and layer. We resample the 3,000 record identifiers with
replacement 10,000 times and recompute the mean paired difference; all
method-specific measurements from a sampled record remain together. The table
reports percentile intervals from the deterministic bootstrap implemented in
the paper asset script.

\input{tables/appendix_paired_bootstrap}

\subsection{Worked Frozen Record and Path Labels}

The main paper traces one unchanged record from direction-fitting statements to
target, neighbor, capability, and path labels. Table~\ref{tab:worked-records-appendix}
adds clean-only, collateral-without-clean, no-effect, and mixed-sign examples.
The mixed example also illustrates why Clean and damage-any are not
complements: one sign can contain a clean operating coefficient while
collateral movement occurs elsewhere on the full path.

\input{tables/appendix_worked_examples}

\section{Additional Prediction Analysis}

Across all cohorts, split types, targets, and both prediction models, adding
$L$ to $B+M$ changes AUROC by $+0.0021$ and AP by $+0.0005$ on average. Adding
the disjoint weak probe $R$ to $B+M$ changes AUROC by $+0.1774$ and AP by
$+0.1821$. Geometry $G$ is outcome-specific within the RFM cohort: it improves
later enhancement and target-any prediction but reduces capability-damage
AUROC, so we do not treat it as a uniformly beneficial feature family.
Table~\ref{tab:geometry-ablation} reports the corresponding RFM-cohort
differences explicitly.

The main paper reports the cross-model summary and the principal CPU controls.
Here we provide the full layer-selection exclusion, weak-response baseline,
grouped-transfer, and RFM-specific geometry results.

\input{tables/cross_model_replication}

\subsection{Measured-Grid Path Topology}

We recompute descriptive topology on all nonzero measured strengths
$|\alpha|\in\{0.1,0.25,0.5\}$. Across learned directions, a target crossing
exists on 7.34\% of paths, a clean coefficient on 6.94\%, and a nonmonotonic
target-or-damage indicator on 10.90\%; the corresponding random rates are
5.98\%, 5.55\%, and 10.65\%. Strict target-first and damage-first patterns are
rare (0.37\% and 0.46\% for learned directions), so they are descriptive rather
than primary prediction targets. Among learned paths with any clean coefficient,
76.3\% contain exactly one measured clean coefficient. These results justify
evaluating multiple strengths, but not a claim that broad continuous clean
windows are common.

\subsection{Layer-Selection Exclusion and Weak-Response Controls}

The 500 records used to select the two primary blocks are a subset of the
3,000-record cohort. Removing them leaves 2,500 records and 30,000
method--block paths. At block 11, all four learned direction families retain
positive paired Target-any and Clean-any differences from random
(Table~\ref{tab:pilot-exclusion-outcomes}). The complete record-held-out random
forest also remains within 0.01 AUROC of its full-cohort estimate on each
principal outcome (Table~\ref{tab:cpu-prediction-controls}). Thus neither the
outcome nor prediction result is driven by reuse of the layer-selection
subset.

The training-free scalar control uses only the signed $|\alpha|=0.1$ response
matched to each later outcome. The $R$-only random forest instead uses the
multivariate low-dose response profile. For target and clean outcomes, this
profile substantially improves over the scalar score; adding base,
method/block, and static-localization features provides a smaller final AUROC
gain. We report prevalence and AP alongside AUROC because the positive path
labels are sparse. The complete feature set is not uniformly best in AP, so
our conclusion concerns the additional ranking information in the structured
weak response rather than universal dominance on every metric.

\input{tables/appendix_pilot_exclusion_outcomes}
\input{tables/appendix_cpu_prediction_controls}

\subsection{Grouped Transfer and Geometry}

\input{tables/appendix_split_macro}
\input{tables/appendix_geometry_ablation}

\begin{figure}[!htb]
    \centering
    \includegraphics[width=0.82\columnwidth]{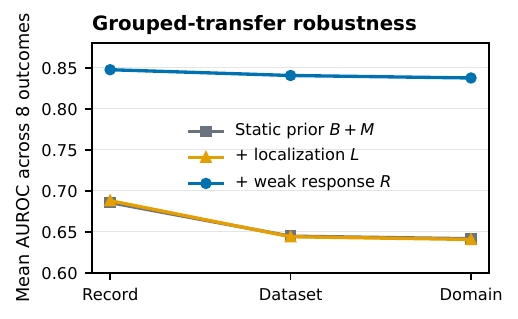}
    \caption{Grouped-transfer robustness. Static localization leaves the
    baseline nearly unchanged, whereas the disjoint weak response produces a
    large AUROC gain under record-, dataset-, and domain-held-out evaluation.}
    \label{fig:split-robustness}
\end{figure}

\section{Multi-Fidelity Strength Selection}

The dense training study contains 500 records, three methods, six layers, and
27 coefficients, yielding 243,000 raw record-strength rows. A disjoint
100-record validation study contributes 24,300 raw dense rows. Excluding the
zero coefficient leaves 234,000 dense training candidates and 23,400 validation
candidates. Multi-fidelity training adds 2,400 non-overlapping records with
sparse trajectories, for 2,900 training records and 320,400 nonzero candidate
rows; validation overlap is zero.

We evaluate selection by held-out policy replay. The global fixed coefficient
is chosen only from mean utility on the 500-record dense training set, which
selects $-1$ for suppression and $+1$ for enhancement. For every unseen
validation path, a learned policy observes metadata, the candidate coefficient,
and the two $|\alpha|=0.1$ weak responses, then either chooses one coefficient
or abstains. Only after this decision do we reveal the measured dense-grid
outcome at the selected coefficient. Thus the validation response curve is not
available to the selector or to fixed-baseline tuning.

Without a weak response, the multi-fidelity $B+M+A$ gradient-boosted decision
tree (GBDT) reaches clean-alpha
AUROC $0.610$ and AP $0.087$. Adding $R$ raises the $M+A+R$ model to AUROC
$0.849$ and AP $0.323$; adding $B$ or $L$ after $R$ does not improve these
ranking metrics consistently. For the same $M+A+R$ model, multi-fidelity
training increases held-out clean rate over dense-only training by 1.0 point
for suppression and 0.8 points for enhancement, while utility changes by less
than $0.001$.

The practical gain comes from risk-aware selection rather than uniformly higher
clean rates (Table~\ref{tab:strength-selector}). No intervention has zero
utility by construction, while the train-tuned fixed $|\alpha|=1$ policy has
negative utility in both directions. The multi-fidelity $M+A+R$ selector abstains on
45.4\% of suppression paths and 39.1\% of enhancement paths, reducing neighbor
damage from 6.0\% to 1.8\% and from 5.2\% to 2.2\%, respectively. Both paired
utility gains over the fixed coefficient are significant. Suppression utility
is also significantly positive relative to no intervention, whereas the
enhancement interval against zero overlaps zero
(Table~\ref{tab:selector-bootstrap}). Its expected cost is 2.55 and 2.61
intervention evaluations per path, about 90\% fewer than the 26-point nonzero dense scan
(Table~\ref{tab:strength-efficiency}). A substantial oracle gap remains, so the
selector is a decision aid rather than a replacement for dense evaluation.

\input{tables/appendix_strength_selector}
\input{tables/appendix_strength_efficiency}
\input{tables/appendix_selector_bootstrap}
\input{tables/appendix_utility_weight_sensitivity}

Selector uncertainty also uses records, not paths, as the sampling unit. We
first average the nine method--layer paths within each of the 100 validation
records, then resample records 10,000 times. Utility and damage comparisons are
paired against the train-tuned fixed policy on the same validation record.
Table~\ref{tab:utility-weight-sensitivity} varies one utility weight at a time
and retrains the utility head. All seven configurations retain positive paired
gains over the fixed policy and reduce neighbor damage, whereas gains over no
intervention are not universal. The declared weights are therefore not the only
configuration that supports risk reduction, but the experiment does not imply
deployment-independent optimality.

\input{tables/appendix_endpoint_validation}
\section{Free-Generation Endpoint Stress Tests}

We keep all free-generation evidence in the supplement because the primary
claims concern random-calibrated answer-margin paths. Two path-conditioned
100-record studies cover four directions, two aligned layers, and coefficients
from $-1$ to $1$. On Qwen3-1.7B, decoded text often changes but learned
directions do not produce a target correction; one isolated random correction
is not evidence of controllability. On Qwen3.5, the unmatched raw-alpha stress
test changes roughly half of nonzero-strength target generations and produces
several learned wrong-to-right transitions, but target damage balances or
exceeds these corrections. The experiment therefore establishes endpoint
reachability at strong dose, not reliable endpoint improvement.

\section{Asset Attribution}
Main-paper Figure 1(a,c) contains explanatory imagery rather than measured PML
activations. Panel (a) adapts Activation Atlas by Carter et al. (CC BY 4.0),
and panel (c) adapts Deep Learning Visuals by David V. Godoy (CC BY 4.0).
Panel (b) and all quantitative figures are generated from the paper workflow.

%% file: tables/appendix_dataset_composition.tex
\begin{table}[t]
\centering
\small
\begin{tabular}{lr}
\toprule
Source dataset & Records \\
\midrule
MMLU-Pro & 1,029 \\
MMLU-Redux 2.0 & 571 \\
AI2 ARC & 350 \\
OpenBookQA & 200 \\
SciQ & 200 \\
LiveBench reasoning & 200 \\
HellaSwag & 150 \\
QASC & 150 \\
LiveBench math & 150 \\
\midrule
Total & 3,000 \\
\bottomrule
\end{tabular}
\caption{Source composition of the frozen 3,000-record benchmark.}
\label{tab:dataset-composition}
\end{table}

%% file: sections/computational_environment_generated.tex
Experiments and paper-facing analyses run on Ubuntu 22.04.5 LTS with an Intel Xeon Gold 6530 CPU (14 allocated cores), 117 GiB RAM, and one NVIDIA GeForce RTX 5090 GPU (driver 580.65.06; CUDA toolkit 13.0). The software environment uses Python 3.10.20, PyTorch 2.12.0+cu132, Transformers 5.8.1, scikit-learn 1.7.2, pandas 2.3.3, NumPy 2.2.5, and SciPy 1.15.3.

%% file: tables/appendix_semantic_audit_sensitivity.tex
\begin{table*}[t]
\centering
\small
\setlength{\tabcolsep}{2.6pt}
\begin{tabular*}{\textwidth}{@{\extracolsep{\fill}}lrrrr@{}}
\toprule
Records retained & Target & Clean & N-dmg. & C-dmg. \\
\midrule
All 3,000 & .815/.357 & .810/.330 & .858/.396 & .856/.382 \\
Exclude fails (2,989) & .818/.369 & .813/.341 & .858/.392 & .855/.384 \\
Exclude non-pass (2,957) & .817/.368 & .813/.344 & .857/.398 & .853/.391 \\
\bottomrule
\end{tabular*}
\caption{Sensitivity to the independent expert semantic audit. The stratified
audit covers 108 records (12 per source): 65 pass, 32 minor issue, and 11 fail.
Cells are complete-predictor AUROC/AP after retaining all records, excluding the
11 fails, or excluding all 43 non-pass records.}
\label{tab:semantic-audit-sensitivity}
\end{table*}

%% file: tables/appendix_paired_bootstrap.tex
\begin{table*}[t]
\centering
\small
\setlength{\tabcolsep}{2.4pt}
\begin{tabular*}{\textwidth}{@{\extracolsep{\fill}}lrrrr@{}}
\toprule
\multicolumn{5}{c}{Block 7} \\
\cmidrule(lr){1-5}
Direction & Target & Clean & N-dmg. & C-dmg. \\
\midrule
Mean difference & $+3.3\,[2.0,4.6]$ & $+3.2\,[1.9,4.4]$ & $-0.3\,[-1.3,0.7]$ & $+0.0\,[-1.1,1.1]$ \\
Linear & $+2.1\,[0.8,3.4]$ & $+2.0\,[0.7,3.2]$ & $+0.6\,[-0.4,1.7]$ & $+0.2\,[-0.9,1.3]$ \\
Logistic & $+2.8\,[1.6,4.1]$ & $+2.8\,[1.5,4.0]$ & $+0.9\,[-0.2,1.9]$ & $+0.8\,[-0.3,1.9]$ \\
RFM/AGOP top-1 & $+3.6\,[2.3,4.9]$ & $+3.4\,[2.1,4.8]$ & $+0.3\,[-0.7,1.3]$ & $+0.8\,[-0.4,1.9]$ \\
\midrule
\multicolumn{5}{c}{Block 11} \\
\cmidrule(lr){1-5}
Mean difference & $+2.5\,[1.2,3.8]$ & $+2.3\,[1.0,3.5]$ & $+0.8\,[-0.2,1.8]$ & $+0.0\,[-1.0,1.0]$ \\
Linear & $+1.4\,[0.2,2.7]$ & $+1.4\,[0.2,2.6]$ & $-0.2\,[-1.2,0.7]$ & $+0.0\,[-1.0,1.0]$ \\
Logistic & $+2.4\,[1.2,3.7]$ & $+2.3\,[1.1,3.6]$ & $+0.4\,[-0.6,1.4]$ & $+0.0\,[-1.0,1.0]$ \\
RFM/AGOP top-1 & $+1.4\,[0.2,2.7]$ & $+1.5\,[0.3,2.7]$ & $+0.5\,[-0.5,1.5]$ & $+0.1\,[-0.9,1.2]$ \\
\bottomrule
\end{tabular*}
\caption{Record-paired percentage-point differences from random in the
frozen 3,000-record study, with 95\% intervals from 10,000 record bootstrap
resamples. Positive target/clean differences are favorable; positive damage
differences are unfavorable.}
\label{tab:paired-bootstrap}
\end{table*}

%% file: tables/appendix_worked_examples.tex
\begin{table*}[t]
\centering
\small
\setlength{\tabcolsep}{3.0pt}
\begin{tabular*}{\textwidth}{@{\extracolsep{\fill}}p{0.13\textwidth}p{0.30\textwidth}p{0.17\textwidth}p{0.20\textwidth}c@{}}
\toprule
Type & Record (source) & Path & Onsets S/E/D & Clean \\
\midrule
Mixed & 13c polarization relaxation time (MMLU-Redux) & Mean diff., 11 & $-.25$/--/$+.50$ & 1 \\
Clean only & ferret beverage soy milk (reasoning) & RFM, 11 & --/$+.50$/-- & 1 \\
Collateral & coal mines energy (OpenBookQA) & RFM, 7 & --/$+.25$/$+.25$ & 0 \\
No effect & acid spill water 01 (AI2 ARC) & Logistic, 7 & --/--/-- & 0 \\
\bottomrule
\end{tabular*}
\caption{Additional frozen-record path examples. Onsets list the first
suppression/enhancement/damage threshold crossings; a dash denotes no crossing.
The examples illustrate label construction rather than quantitative evidence.}
\label{tab:worked-records-appendix}
\end{table*}

%% file: tables/cross_model_replication.tex
\begin{table*}[t]
\centering
\small
\setlength{\tabcolsep}{1.5pt}
\begin{tabular*}{\textwidth}{@{\extracolsep{\fill}}lrrrrrr@{}}
\toprule
Cohort & $B{+}M$ & $\Delta L$ & $\Delta R$ & Final & Dataset & Domain \\
\midrule
Qwen3-1.7B primary & 0.686 & +0.2 & \textbf{+16.1} & \textbf{0.849} & 0.843 & 0.840 \\
Qwen3-1.7B subset & 0.612 & -0.1 & \textbf{+21.7} & \textbf{0.828} & 0.829 & 0.828 \\
Qwen3.5-2B & 0.621 & +1.3 & \textbf{+17.1} & \textbf{0.805} & 0.806 & 0.800 \\
Ministral-3B & 0.615 & +2.2 & \textbf{+16.4} & \textbf{0.801} & 0.796 & 0.800 \\
\bottomrule
\end{tabular*}
\caption{Macro AUROC across the eight outcomes visualized in the main paper. $\Delta L$ and $\Delta R$ are cumulative gains; Final is $B+M+L+R$. Dataset and Domain use the final predictor. The Qwen3-1.7B subset comes from the primary cohort.}
\label{tab:cross-model-prediction}
\end{table*}

%% file: tables/appendix_pilot_exclusion_outcomes.tex
\begin{table}[t]
\centering
\small
\setlength{\tabcolsep}{3.6pt}
\begin{tabular}{@{}lrr@{}}
\toprule
Direction & Target-any & Clean-any \\
\midrule
Mean difference & .034 [.021,.048] & .033 [.019,.047] \\
RFM/AGOP & .034 [.020,.048] & .032 [.019,.046] \\
Logistic & .026 [.012,.040] & .024 [.011,.038] \\
Linear & .016 [.003,.029] & .016 [.003,.028] \\
\bottomrule
\end{tabular}
\caption{Sensitivity after excluding the 500-record layer-selection subset.
Entries are learned-minus-random path-incidence differences at block 11 with
record-paired 95\% bootstrap intervals on the remaining 2,500 records.}
\label{tab:pilot-exclusion-outcomes}
\end{table}

%% file: tables/appendix_cpu_prediction_controls.tex
\begin{table}[t]
\centering
\small
\setlength{\tabcolsep}{1.0pt}
\begin{tabular*}{\columnwidth}{@{\extracolsep{\fill}}lrrrr@{}}
\toprule
\multicolumn{5}{c}{(a) Scalar and multivariate weak-response controls} \\
\cmidrule(lr){1-5}
Outcome & Prev. & Scalar & $R$-only & Full \\
\midrule
Supp. & .058 & .793/.296 & .843/.338 & .865/.327 \\
Enh. & .056 & .771/.282 & .833/.320 & .852/.312 \\
Target-any & .111 & .708/.324 & .793/.367 & .815/.357 \\
Neighbor dmg. & .079 & .840/.408 & .842/.402 & .858/.396 \\
Capability dmg. & .078 & .849/.373 & .842/.373 & .856/.382 \\
Clean supp. & .054 & .729/.227 & .841/.308 & .863/.299 \\
Clean enh. & .053 & .735/.239 & .832/.305 & .851/.292 \\
Clean-any & .105 & .651/.266 & .789/.342 & .810/.330 \\
\bottomrule
\end{tabular*}

\begin{tabular*}{\columnwidth}{@{\extracolsep{\fill}}lrrr@{}}
\toprule
\multicolumn{4}{c}{(b) Complete random forest after subset exclusion} \\
\cmidrule(lr){1-4}
Outcome & Prev. & AUROC [95\% CI] & AP [95\% CI] \\
\midrule
Target-any & .101 & .824 [.810,.838] & .355 [.298,.412] \\
Clean-any & .095 & .819 [.806,.831] & .328 [.278,.377] \\
Neighbor dmg. & .075 & .858 [.848,.868] & .392 [.369,.414] \\
Capability dmg. & .078 & .852 [.850,.856] & .373 [.338,.400] \\
\bottomrule
\end{tabular*}
\caption{Record-held-out CPU prediction controls on Qwen3-1.7B-Base. In
panel (a), each cell is AUROC/AP on all 3,000 records. Scalar is the
training-free signed weak-response score; $R$-only and Full are random forests,
where Full uses $B+M+L+R$. Panel (b) reports prevalence, AUROC, and AP with
95\% intervals after excluding the 500-record layer-selection subset.}
\label{tab:cpu-prediction-controls}
\end{table}

%% file: tables/appendix_split_macro.tex
\begin{center}
\centering
\small
\begin{tabular}{llrr}
\toprule
Split & Features & AUROC & AP \\
\midrule
Record & $B+M$ & .686 & .144 \\
Record & $B+M+L$ & .688 & .145 \\
Record & $B+M+R$ & .848 & .333 \\
Dataset & $B+M$ & .645 & .129 \\
Dataset & $B+M+L$ & .644 & .126 \\
Dataset & $B+M+R$ & .841 & .305 \\
Domain & $B+M$ & .642 & .125 \\
Domain & $B+M+L$ & .641 & .123 \\
Domain & $B+M+R$ & .838 & .303 \\
\bottomrule
\end{tabular}
\captionof{table}{Macro performance over the eight confirmatory later-strength outcomes. Values are
random-forest AUROC/AP in the all-method cohort.}
\label{tab:split-macro}
\end{center}

%% file: tables/appendix_geometry_ablation.tex
\begin{center}
\centering
\small
\begin{tabular}{lrrrr}
\toprule
Outcome & Base & $+G$ & $\Delta$ AUC & $\Delta$ AP \\
\midrule
Suppression & .651 & .646 & $-.005$ & $-.007$ \\
Enhancement & .645 & .680 & $+.035$ & $+.005$ \\
Target-any & .666 & .688 & $+.022$ & $+.014$ \\
Neighbor damage & .632 & .635 & $+.003$ & $+.008$ \\
Capability damage & .622 & .589 & $-.033$ & $-.021$ \\
Clean suppression & .652 & .649 & $-.002$ & $-.003$ \\
Clean enhancement & .644 & .680 & $+.037$ & $-.002$ \\
Clean-any & .666 & .690 & $+.025$ & $+.018$ \\
\bottomrule
\end{tabular}
\captionof{table}{RFM-cohort geometry ablation, averaged over record-, dataset-, and
domain-grouped splits. $G$ contains AGOP spectrum and alignment features.
Deltas compare $B+M+L+G$ with the $B+M+L$ base.}
\label{tab:geometry-ablation}
\end{center}

%% file: tables/appendix_strength_selector.tex
\begin{table}[t]
\centering
\small
\setlength{\tabcolsep}{3pt}
\begin{tabular}{@{}lrrrr@{}}
\toprule
\multicolumn{5}{c}{Suppression} \\
\cmidrule(lr){1-5}
Selector & Clean & Utility & N-dmg. & Abstain \\
\midrule
No intervention & .000 & .000 & .000 & 1.000 \\
Fixed $|\alpha|=1$ & .098 & $-.041$ & .060 & .000 \\
Dense $M+A+R$ & .106 & .013 & .016 & .473 \\
MF $B+M+A$ & .060 & $-.018$ & .037 & .447 \\
MF $M+A+R$ & .116 & .013 & .018 & .454 \\
MF $B+M+A+R$ & .110 & .012 & .018 & .483 \\
MF $B+M+L+A+R$ & .098 & .011 & .019 & .471 \\
Dense oracle & .232 & .084 & -- & -- \\
\bottomrule
\end{tabular}

\begin{tabular}{@{}lrrrr@{}}
\toprule
\multicolumn{5}{c}{Enhancement} \\
\cmidrule(lr){1-5}
Selector & Clean & Utility & N-dmg. & Abstain \\
\midrule
No intervention & .000 & .000 & .000 & 1.000 \\
Fixed $|\alpha|=1$ & .119 & $-.026$ & .052 & .000 \\
Dense $M+A+R$ & .109 & .008 & .024 & .407 \\
MF $B+M+A$ & .079 & $-.017$ & .042 & .312 \\
MF $M+A+R$ & .117 & .008 & .022 & .391 \\
MF $B+M+A+R$ & .112 & .010 & .024 & .422 \\
MF $B+M+L+A+R$ & .113 & .010 & .026 & .417 \\
Dense oracle & .260 & .089 & -- & -- \\
\bottomrule
\end{tabular}
\caption{Held-out dense-grid strength selection on 100 records.
$A$ denotes candidate-strength features; MF denotes multi-fidelity training.}
\label{tab:strength-selector}
\end{table}

%% file: tables/appendix_strength_efficiency.tex
\begin{table}[t]
\centering
\small
\begin{tabular}{lrr}
\toprule
Policy & Suppression & Enhancement \\
\midrule
Train-tuned fixed & 1.000 & 1.000 \\
Dense-only $M+A+R$ & 2.527 & 2.593 \\
Multi-fidelity $M+A+R$ & 2.546 & 2.609 \\
Dense scan & 26.000 & 26.000 \\
\midrule
Multi-fidelity reduction & 90.2\% & 90.0\% \\
\bottomrule
\end{tabular}
\caption{Expected intervention evaluations per held-out path. Learned policies
use two weak probes and execute the selected coefficient only when they do not
abstain. Reduction is relative to evaluating all 26 nonzero dense-grid
coefficients.}
\label{tab:strength-efficiency}
\end{table}

%% file: tables/appendix_selector_bootstrap.tex
\begin{table}[t]
\centering
\small
\setlength{\tabcolsep}{2.8pt}
\begin{tabular*}{\columnwidth}{@{\extracolsep{\fill}}lrr@{}}
\toprule
Metric & Suppression & Enhancement \\
\midrule
Utility vs. no intervention & .013 [.003,.025] & .008 [$-.001$,.019] \\
Utility vs. fixed & .055 [.044,.066] & .034 [.024,.044] \\
Target change & 0.9 [$-1.3$,3.0] & $-0.8$ [$-2.6$,1.1] \\
Clean change & 1.8 [0.0,3.8] & $-0.2$ [$-1.9$,1.8] \\
Neighbor-dmg. change & $-4.2$ [$-6.3$,$-2.3$] & $-3.0$ [$-4.9$,$-1.4$] \\
Capability-dmg. change & $-1.1$ [$-2.0$,$-0.4$] & $-0.2$ [$-0.7$,0.2] \\
\bottomrule
\end{tabular*}
\caption{Record-paired uncertainty for the multi-fidelity $M+A+R$
selector on 100 held-out records. Brackets are 95\% intervals from 10,000
record bootstrap resamples. Outcome changes are selector-minus-fixed in
percentage points.}
\label{tab:selector-bootstrap}
\end{table}

%% file: tables/appendix_utility_weight_sensitivity.tex
\begin{table}[t]
\centering
\small
\setlength{\tabcolsep}{3.0pt}
\begin{tabular}{@{}lrr@{}}
\toprule
Configuration & $\Delta U$ vs. fixed & $\Delta U$ vs. zero \\
\midrule
Damage weight 0.5 & .030/.015 & .021/.020 \\
No clean bonus & .056/.041 & .015/.015 \\
No strength penalty & .050/.032 & .009/.006 \\
Declared weights & .055/.034 & .013/.008 \\
Clean bonus 0.2 & .041/.026 & .000/.000 \\
Strength penalty 0.02 & .056/.037 & .015/.011 \\
Damage weight 2.0 & .113/.090 & .008/.003 \\
\bottomrule
\end{tabular}
\caption{Selector utility sensitivity on the same 100 held-out records. Each
row retrains the utility head after changing one declared weight. Entries are
paired utility gains for suppression/enhancement.}
\label{tab:utility-weight-sensitivity}
\end{table}

%% file: tables/appendix_endpoint_validation.tex
\begin{table}[t]
\centering
\small
\setlength{\tabcolsep}{1.8pt}
\begin{tabular*}{\columnwidth}{@{\extracolsep{\fill}}lrrrrr@{}}
\toprule
\multicolumn{6}{c}{Qwen3-1.7B} \\
\cmidrule(lr){1-6}
Direction & Paths & Base & G/D & Text $\Delta$ & Clause $\Delta$ \\
\midrule
Random & 24 & 5.6\% & 1/0 & 37.2\% & 21.5\% \\
Mean diff. & 24 & 2.8\% & 0/1 & 31.9\% & 19.6\% \\
Logistic & 28 & 4.8\% & 0/0 & 25.4\% & 12.9\% \\
RFM/AGOP & 24 & 6.9\% & 0/1 & 33.2\% & 13.2\% \\
\midrule
\multicolumn{6}{c}{Qwen3.5-2B} \\
\cmidrule(lr){1-6}
Random & 24 & 8.3\% & 0/0 & 45.0\% & 19.4\% \\
Mean diff. & 24 & 11.1\% & 2/4 & 50.3\% & 30.4\% \\
Logistic & 28 & 8.3\% & 3/2 & 52.1\% & 32.6\% \\
RFM/AGOP & 24 & 8.3\% & 2/4 & 47.4\% & 28.5\% \\
\bottomrule
\end{tabular*}
\caption{Path-conditioned free-generation endpoint stress tests on 100 records
per model. G/D counts unique target prompts with correctness gain/damage at any
nonzero coefficient. Text and clause changes are measured against $\alpha=0$.
Qwen3.5 uses the unmatched raw-alpha stress protocol.}
\label{tab:endpoint-validation}
\end{table}